\pdfoutput=1

\documentclass[11pt,dvipsnames]{article}

\usepackage[]{acl}

\usepackage{times}
\usepackage{latexsym}

\usepackage[T1]{fontenc}

\usepackage[utf8]{inputenc}

\usepackage{microtype}
\usepackage{enumitem}

\usepackage{inconsolata}

\usepackage{graphicx}

\usepackage{multirow}
\usepackage{booktabs}
\usepackage{graphicx}
\usepackage{float}
\usepackage{algorithm}
\usepackage{algpseudocode}
\usepackage{caption}
\usepackage{subfigure}
\usepackage{amssymb}
\usepackage{soul}

\usepackage[utf8]{inputenc} 
\usepackage[T1]{fontenc}    
\usepackage{hyperref}       
\usepackage{url}            
\usepackage{wrapfig,booktabs}       
\usepackage{amsfonts}       
\usepackage{nicefrac}       
\usepackage{microtype}      
\usepackage{xcolor}
\usepackage{amsmath}
\usepackage{amssymb}
\usepackage{pgf-pie}
\usepackage{graphicx}
\usepackage{pgfplots}
\usetikzlibrary{patterns}
\pgfplotsset{compat=1.18}
\usepackage{graphicx}
\usepackage{pythonhighlight}
\usepackage{bbding}
\usepackage{listings}
\usepackage{geometry}
\usepackage{array}
\usepackage{fvextra}
\usepackage{listings}

\usepackage{tikz}
\usepackage{fontawesome5}

\newcommand{\colorLetter}[2]{\textcolor{#1}{#2}}

\newcommand\model{\textsc{\textbf{{\colorLetter{Red}{E}\colorLetter{Blue}{m}\colorLetter{ForestGreen}{m}\colorLetter{Orange}{a}-\colorLetter{Purple}{X}}}}}

\newcommand{\SR}{\textsc{Spatial Relation}}
\newcommand{\OODO}{\textsc{OOD Object}}
\newcommand{\OODI}{\textsc{OOD Instruction}}
\newcommand{\ID}{\textsc{In Domain}}

\definecolor{githubdark}{RGB}{36, 41, 46}     
\definecolor{hugfaceyellow}{RGB}{255, 193, 7} 
\definecolor{webblue}{RGB}{41, 128, 185}      
\hypersetup{
    colorlinks=true,
    linkcolor=blue,
    filecolor=magenta,      
    urlcolor=black,
}

\title{\model{}: An \colorLetter{Red}{E}mbodied \colorLetter{Blue}{M}ulti\colorLetter{ForestGreen}{m}odal \colorLetter{Orange}{A}ction \colorLetter{Purple}{Model} with\\ Grounded Chain of Thought and Look-ahead Spatial Reasoning}

\author{Qi Sun$^{1}$\textsuperscript{*}, Pengfei Hong$^{1}$\textsuperscript{*}, Tej Deep Pala$^{1}$\vphantom{\thanks{Both authors contributed equally to this work. The first authorship was randomly assigned by coin flip.}}, Vernon Y.H. Toh$^{1}$,\\ \textbf{U-Xuan Tan$^{1}$, Deepanway Ghosal$^1$\thanks{Now at Deepmind.}, Soujanya Poria$^1$} \\\\
$^1$ Singapore University of Technology and Design
}
\makeatletter
\AtBeginDocument{
\let\@oldmaketitle\@maketitle
\renewcommand{\@maketitle}{\@oldmaketitle
  \vspace{-27pt}
  \begin{center}
  \includegraphics[width=\linewidth]{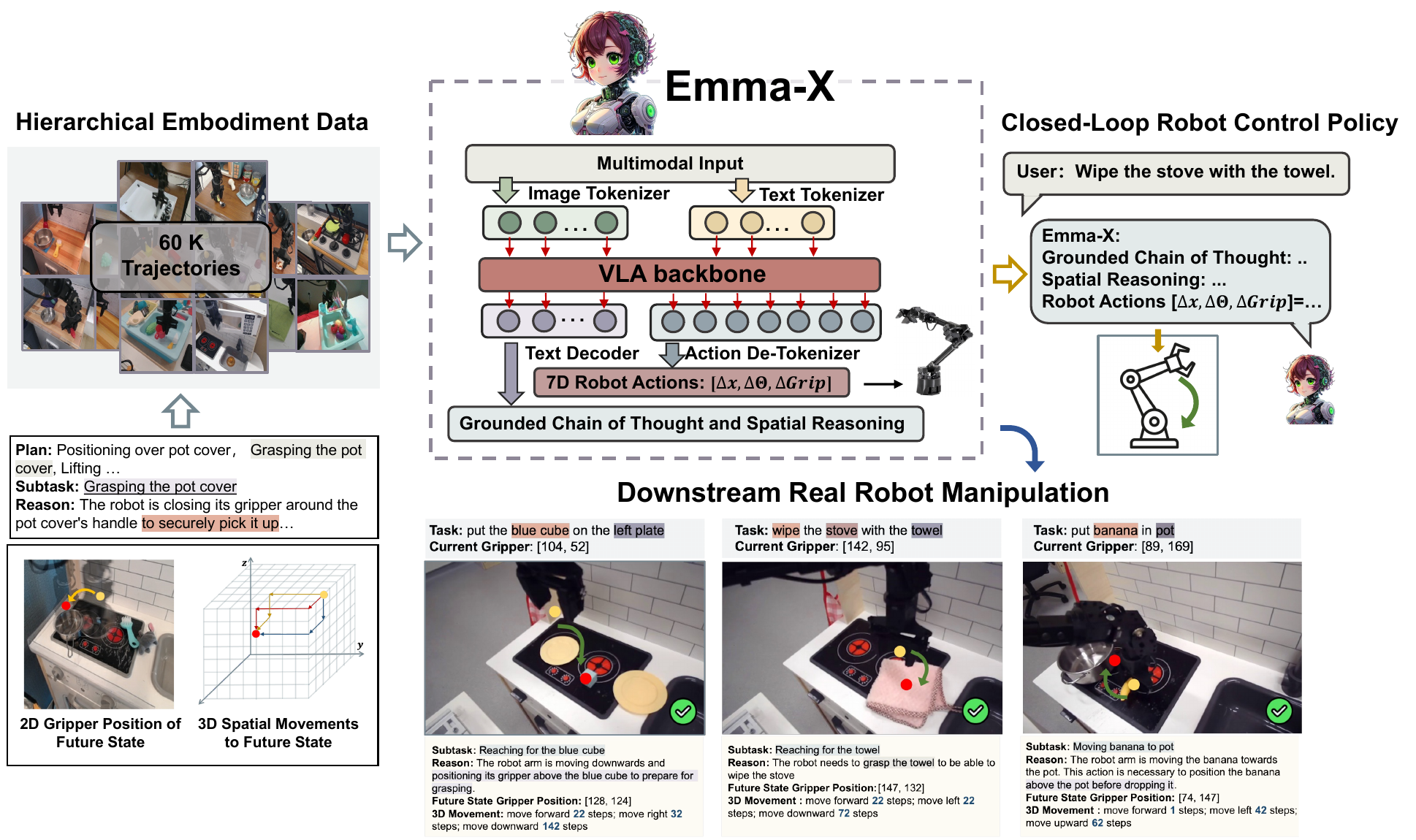}
    \href{https://github.com/declare-lab/Emma-X}{
        {\color{githubdark}{\scalebox{2}{\faGithub}}} 
    } \hspace{0.1cm}
    \href{https://huggingface.co/declare-lab/Emma-X}{
        {\color{hugfaceyellow}{\scalebox{2}{\faRobot}}} 
    } \hspace{0.1cm}
    \href{https://declare-lab.github.io/Emma-X/}{
        {\color{webblue}{\scalebox{2}{\faGlobe}}} 
    }
  \end{center}  
 }
}
\makeatother
\begin{document}
\maketitle
\begin{tikzpicture}[remember picture,overlay,shift={(current page.north west)}]
\node[anchor=north west,xshift=0.5cm,yshift=-1.7cm]{\scalebox{1}[1]{\includegraphics[width=3.5cm]{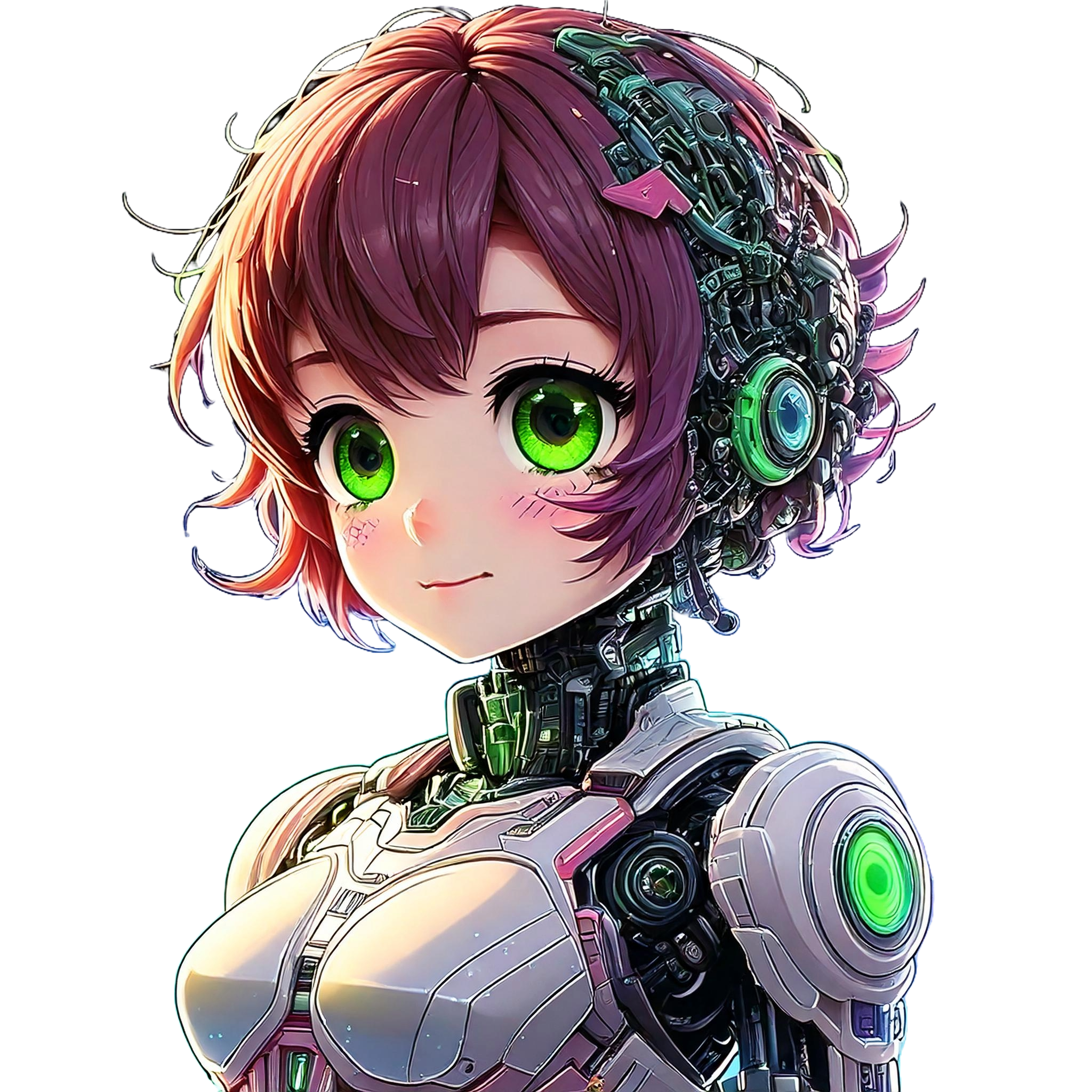}}};
\end{tikzpicture}
\begin{abstract}
Traditional reinforcement learning-based robotic control methods are often task-specific and fail to generalize across diverse environments or unseen objects and instructions. Visual Language Models (VLMs) demonstrate strong scene understanding and planning capabilities but lack the ability to generate actionable policies tailored to specific robotic embodiments.  To address this, Visual-Language-Action (VLA) models have emerged, yet they face challenges in long-horizon spatial reasoning and grounded task planning. In this work, we propose the Embodied Multimodal Action Model with Grounded Chain of Thought and Look-ahead Spatial Reasoning, \model{}. \model{} leverages our constructed hierarchical embodiment dataset based on BridgeV2, containing 60,000 robot manipulation trajectories auto-annotated with grounded task reasoning and spatial guidance. Additionally, we introduce a trajectory segmentation strategy based on gripper states and motion trajectories, which can help mitigate hallucination in grounding subtask reasoning generation. Experimental results demonstrate that \model{} achieves superior performance over competitive baselines, particularly in real-world robotic tasks requiring spatial reasoning. We make our codes, models and datasets publicly available: \url{https://declare-lab.github.io/Emma-X/}.
\end{abstract}

\section{Introduction}
The robotic policy model aims to generate sequences of low-level action manipulation policies for robots. Traditional reinforcement learning-based robotic control methods often focus on narrowly defined tasks within fixed environments \citep{ma2024survey}, hindering their ability to generalize beyond task-specific training data and limiting their applicability \citep{brohan2023rt1roboticstransformerrealworld,chi2023diffusion}. 

Recent advancements in foundation models for vision and language have highlighted the remarkable scene-understanding and task-planning capabilities \citep{radford2021learning,zhai2023sigmoid,touvron2023llama}. These Visual-Language Models (VLMs) excel at breaking down complex tasks into manageable steps through chain-of-thought reasoning and demonstrate significant potential in planning. Despite their strengths, VLMs are not inherently designed to directly generate policies applicable to specific embodiment configurations in robotics. This limitation has spurred the emergence of Visual-Language-Action (VLA) models, which aim to bridge this gap by leveraging multimodal inputs to produce adaptive and generalized robotic actions for complex, multi-task scenarios \citep{brohan2023rt2visionlanguageactionmodelstransfer,kim2024openvla,octo_2023}. 

\begin{figure}
	\centering
	\includegraphics[width=\linewidth]{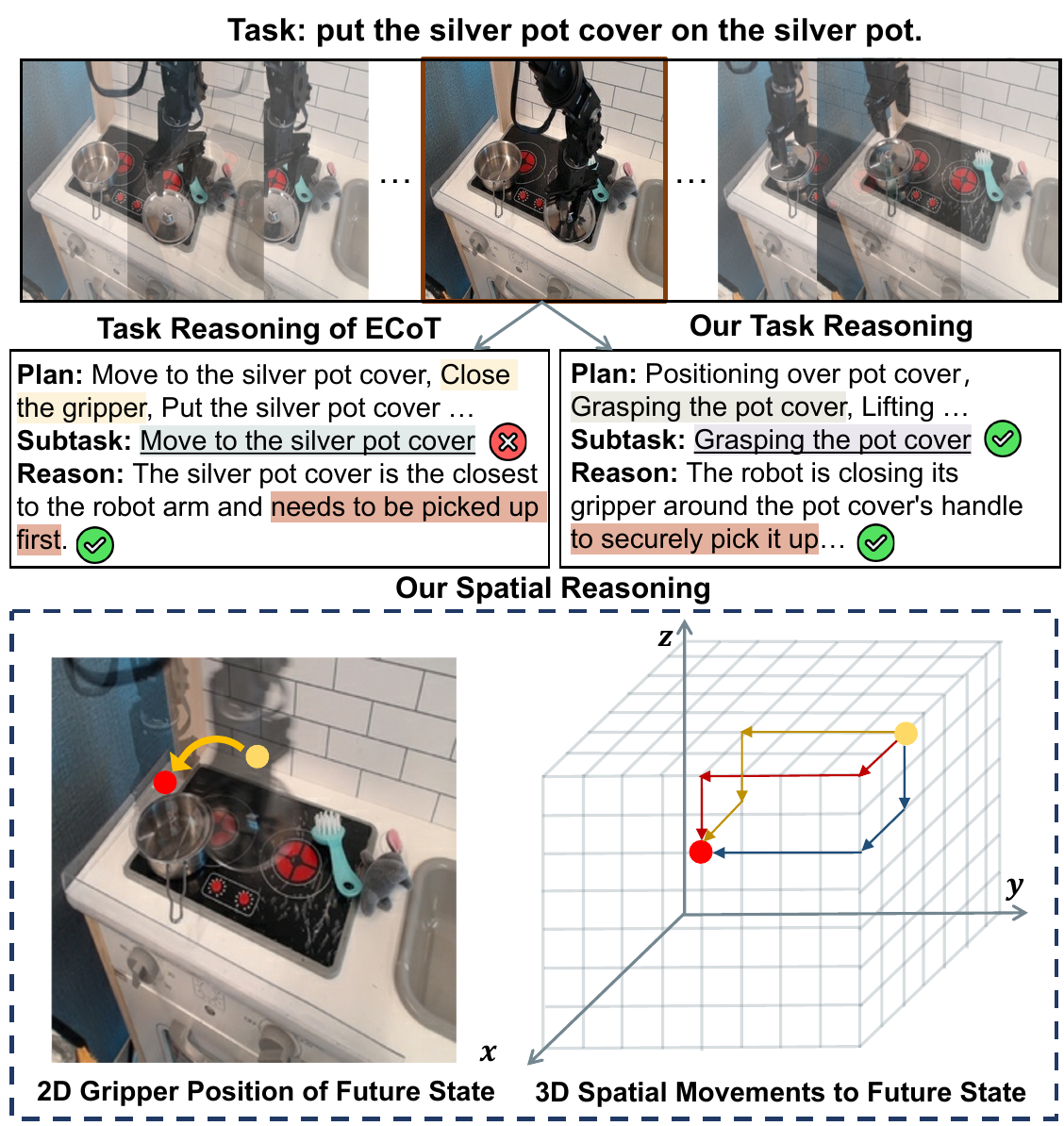}
	\caption{Comparison of our \model \, with ECoT in task reasoning. While both approaches utilize Gemini, our method also incorporates image sequence input, whereas ECoT relies solely on text input. We also illustrate an example of spatial reasoning.}
	\label{fig:intro} 
     \vspace{-0.9em}

\end{figure}

However, most of the existing VLA models often exhibit ``muscle memory'' response patterns, struggling to perceive scene variation and understand instructions as humans do when handling complex tasks or ambiguous commands. \citet{zawalski2024robotic} attempts to address this issue through visual and task reasoning, including the bounding box of the object, task segmentation, and the direction of predicted action, etc.  Although they equip VLAs with an understanding of the current situation and task, they lack long-horizon spatial reasoning on how robots should move next. We hypothesize that the completion of subgoals or subtasks can be enhanced if the VLA incorporates look-ahead spatial reasoning, such as inferring the gripper's future 2D position and the 3D movement plans necessary for the gripper to reach that position. In particular, we train the VLA model to predict future position \( g_{t+k} \) of the gripper as checkpoints and use them to devise a high-level movement plan \( \beta(g_t, g_{t+k}) \). This plan informs the immediate action \( a_t \) at the current state \( s_t \), ensuring decisions are both reactive to the present and aligned with long-term objectives. Similar to a delivery driver planning a route with key landmarks to make purposeful driving decisions, this approach optimizes task completion by balancing foresight and adaptability. 

Additionally, another limitation in task reasoning provided by ECoT\citep{zawalski2024robotic} is the absence of visual grounding when augmenting reasoning data using Gemini. We observe that Gemini frequently hallucinates due to a lack of holistic understanding of the setup and environment. As shown in Figure \ref{fig:intro}, the image shows that the robot already started to grasp the pot cover, while the task reasoning indicates the subtask is still ``Move to silver pot cover", which conflicts with the following reasoning they provided. 

In this work, we introduce the Embodied Multimodal Action Model with Grounded Chain of Thought Reasoning, \model. We develop a hierarchical embodiment dataset based on BridgeV2, consisting of 60,000 robot manipulation trajectories. For each state of a given trajectory, we generate detailed spatial reasoning grounded in the environment and task reasoning, such as the plans of how the robot should perform the subtask. As shown in Figure \ref{fig:intro}, we also generate the 2D gripper position, and 3D spatial movements of the gripper to transit to future states, which enable the VLA model to reason a long-horizon plan for accomplishing the task.

Furthermore, we utilize Gemini \citep{team2023gemini} to generate grounded task reasoning for each observed state. To avoid the abovementioned reasoning conflict problem of task reasoning in ECoT, we propose a novel trajectory segmentation strategy, which leverages the opening and closing states of the gripper and the motion trajectory of the robot arm to segment the sequence of states into distinct segments. By grounding, we mean that, unlike ECoT, which prompts Gemini to generate subtask reasoning based solely on textual descriptions, our approach incorporates visual images segmented using the aforementioned strategy. As shown in Figure \ref{fig:intro}, our method can accurately provide the subtask ``Grasping the pot cover" corresponding to the current robotic state. This illustrates that our strategy significantly reduces Gemini's hallucination issues by requiring it to construct a visual understanding of the environment, rather than relying solely on textual descriptions of the environment. Finally, we train our \model \, based on OpenVLA using our constructed hierarchical embodiment dataset.

The main contributions of our work are summarized as follows:
\begin{itemize}[noitemsep]
\item We introduce a 7B-parameter embodied multimodal action model, \model{} created by fine-tuning OpenVLA with the grounded chain of thought (CoT) reasoning data.
\item We synthetically construct a hierarchical embodiment dataset from the existing robot manipulation dataset, which includes the 3D spatial movements, 2D gripper position, and grounded reasoning.
\item We propose a novel trajectory segmentation strategy that leverages the gripper's opening and closing states alongside the motion trajectory of the robot arm, facilitating both grounded task reasoning and look-ahead spatial reasoning.

\item Our proposed \model \, achieves significant performance improvements over existing competitive baselines on various real-world robot tasks, especially in tasks where spatial reasoning is required.

\end{itemize}

\section{Problem Formulation}
\subsection{Policy Imitation Learning}
Given a set of expert demonstrations \( \mathcal{D} = \{(\{s_t\}_{t=1}^{T}, \mathcal{T}_i, \{a_t\}_{t=1}^{T})\}_{i=1}^{N} \), where \(N\) is the number of demonstrations in the dataset, \(T\) is the number of states (image frames of the environment) for a data sample \(D_i\), \( s_i = \text{image}_i \) represents the state consisting of an image of the environment, \(\mathcal{T}_i\) is a natural language task instruction, and \( a_i \) represents the action taken by the expert in that state, the goal is to learn a policy \( \pi_{\theta}(a \mid s, \mathcal{T}) \) that mimics the expert's behavior.

The policy $\pi_{\theta}$ is modeled by a Vision-Language-Action (VLA) model. In line with the OpenVLA setting, the policy outputs a generalized action as a 7-dimensional vector. This vector encodes the end-effector's (gripper's)  velocity of Cartesian components $(x, y, z)$, orientational components (roll, pitch, yaw), and the gripper's close-open action.


The goal is to find parameters \( \theta \) that minimize the difference between predicted action and the expert's action. 

\subsection{Hierarchical Policy Imitation}
\label{ss:hierachical_policy_imitation}
We build on the above formulation by decomposing a general task \( \mathcal{T} \) into a hierarchical structure consisting of finer-grained components: states, segments, and subtasks.

A \textbf{state} at timestep \( t \), denoted \( s_t \), represents the scene. The sequence of states for the \( i \)-th trajectory is \( S_i = \{s_1, s_2, \ldots, s_T\} \), where \( T \) is the number of timesteps. An \textbf{action} \( a_t \) is taken at state \( s_t \), and the corresponding sequence of actions is \( A_i = \{a_1, a_2, \ldots, a_T\} \). A \textbf{segment} \( \sigma \) is a series of consecutive states, \( \{s_t, s_{t+1}, \ldots, s_{t+k}\} \), contributing to a subgoal, with \( \Sigma_i = \{\sigma_1, \sigma_2, \ldots, \sigma_n\} \) representing the segment sequence for the \( i \)-th trajectory. In each segment, the robot performs similar actions. A \textbf{subtask} \( \mathcal{S} \) consists of segments, \( \{\sigma_1, \sigma_2, \ldots, \sigma_p\} \), to achieve a specific subgoal. Finally, a \textbf{task} \( \mathcal{T} \) is a series of subtasks, \( \{\mathcal{S}_1, \mathcal{S}_2, \ldots, \mathcal{S}_m\} \), required to complete the overall objective.



Our Vision-Language-Action (VLA) model \( \pi_{\theta}(a_t \mid s_t, \mathcal{T}) \) predicts actions \( a_t \) for each state \( s_t \) by hierarchically decomposing tasks into subtasks. This ensures the end-effector's motion aligns with subgoal intents, enhancing the model's ability to execute complex tasks through manageable subtasks. We create a dataset \( \mathcal{D} = \{\mathcal{D}_i\}_{i=1}^{N} \), where \( \mathcal{D}_i = \{S_i, \Sigma_i, \mathcal{T}_i\} \). Each state \( s_t \in S_i \) is labeled with its subtask. Without such labeling, chain-of-thought training is infeasible. During inference, the model generates reasoning chains, including subtasks and relevant spatial information derived from visual scenes.
 
\begin{figure*}
	\centering
	\includegraphics[width=\textwidth]{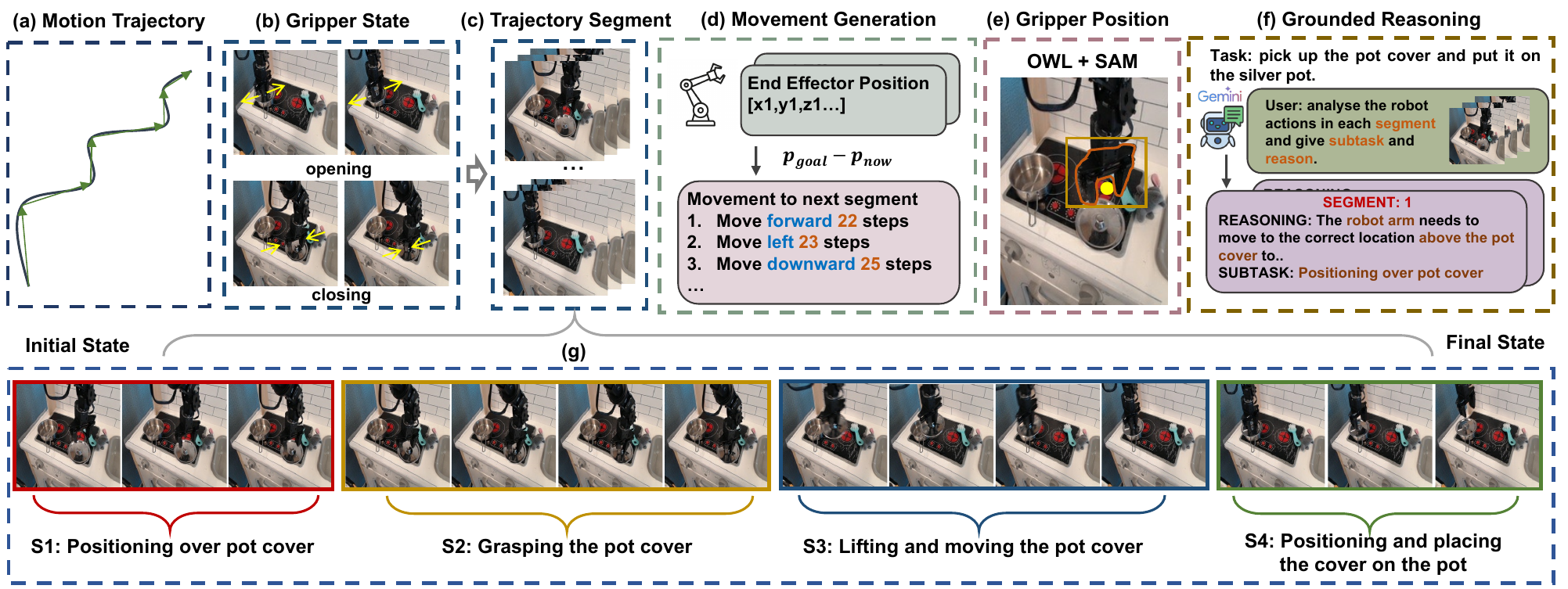}
	\caption{Construction of our hierarchical embodied dataset. We first segment the trajectory. Then, we generate the 3D spatial movement that requires to transition to the end state of the segment. Based on segments, we recognize the 2D gripper position and generate the grounded task reasoning.}
	\label{fig:data}
     \vspace{-0.8em}

\end{figure*}

\section{Methodology}
In this section, we introduce our proposed framework in detail. Our \model{} encompasses three crucial designs: (1) Segmenting the trajectory based on the states of the gripper and the motion trajectory of the robotic arm. (2) Generating hierarchical planning including grounded task reasoning, 2D gripper positions, and 3D spatial movements. (3) Training the our \model{} based on OpenVLA with our constructed dataset.

\subsection{Trajectory Segmentation}
\label{sec:segment}

\paragraph{Why Segment Trajectories?}  
The overarching goal of our work is to enhance Vision-Language-Action (VLA) models with grounded chain-of-thought (CoT) reasoning. We identified two key limitations in existing VLAs:  1) While existing VLAs improve task decomposition by breaking a task into subtasks and solving each using CoT~\cite{zawalski2024robotic}, their CoT reasoning relies exclusively on textual scene descriptions~\footnote{We use ``scene'' and ``environment'' interchangeably throughout this paper.}. This limits their reasoning capability for real-world scenarios.  2) They lack robust spatial reasoning abilities, essential for effective task planning and execution.  

To address these limitations, we propose two key solutions:  \textbf{Incorporating visual scene information:} Beyond textual prompts, we integrate visual inputs into Gemini to enable task decomposition into subtasks and generate high-level plans grounded in both visual and textual contexts. \textbf{Fine-grained movement plans:} We train the robot to determine \emph{where to go} and \emph{how to reach} a potential future state necessary for completing a subtask.  

To implement these solutions, every state must be labeled with the subtask the robot is performing. However, our experiments revealed that directly annotating each individual frame via Gemini resulted in noisy labels, likely due to insufficient contextual information. To overcome this, we segment trajectories into sequences of consecutive states where the robot performs semantically similar actions. This segmentation provides richer context, allowing Gemini to assign subtask labels more effectively.  

Additionally, segmentation facilitates finding the gripper's position in a future state and planning its movement. At a given state \( s_t \), the model predicts the movement plan required to reach the initial state of the next segment, \( s_{t+k} \), before determining the policy \( a_t \) for \( s_t \). Since \( t+k > t \), this approach enables the model to perform look-ahead spatial reasoning, predicting the gripper's position at a likely future state, planning the motion trajectory, and generating \( a_t \) accordingly.  

\paragraph{Our Segmentation Method.}
As shown in Figure~\ref{fig:data}(a) and Figure~\ref{fig:data}(b), we segment observation sequences by integrating the motion trajectory and the gripper states of the end effector. To achieve this, we utilize the \textbf{Hierarchical Density-Based Spatial Clustering of Applications with Noise (HDBSCAN)} algorithm \citep{McInnes2017hdbscanHD}, which effectively handles noise stemming from small fluctuations caused by imperfections in human demonstration. The flexibility of HDBSCAN enables the discovery of diverse trajectory patterns within the data.

We define a custom distance measurement to segment the end effector's trajectory, capturing both spatial and temporal information. Let \( \mathbf{p}_i = (x_i, y_i, z_i) \) denote the 3D position, and \( \mathbf{r}_i = (r_{ix}, r_{iy}, r_{iz}) \) represent the 3D orientation of data point \( i \). Additionally, let \( t_i \) represent the timestamp of this data point. The distance between two data points \( i \) and \( j \) is given by the following expression:
\begin{equation}  
d(i, j) = \|\mathbf{p}_i - \mathbf{p}_j\|_2 + \lambda \|\mathbf{r}_i - \mathbf{r}_j\|_2 + \beta |t_i - t_j|
\label{eq:distance-measurement}
\end{equation}
where \( \lambda \) is a weighting factor for the orientation component, and \( \beta \) controls the influence of the temporal distance~\footnote{We use \(\lambda\) as 1 and \(\beta\) as 0.03 for best segmentation.}. This combined distance metric \( d(i, j) \) ensures that both spatial movement and temporal separation contribute to the segmentation process. The inclusion of temporal information helps to distinguish trajectories that are spatially similar but occur at different times, while the orientation term captures changes in the end effector's rotation.

Applying the HDBSCAN algorithm with this distance metric allows us to segment the trajectory into meaningful clusters that reflect distinct movement patterns. However, the motion trajectory alone does not fully capture the interaction dynamics of the end effector with the environment. To address this, we incorporate the gripper state \( gs_i \), which represents whether the gripper is in a \textit{grip} (closed) or \textit{loose} (open) position. A segmentation breakpoint occurs when the HDBSCAN algorithm detects a new cluster or when the gripper state changes between consecutive data points, formally defined as \( gs_i \neq gs_{i+1} \).

This dual-segmentation approach effectively combines trajectory-based clustering with interaction-based segmentation, ensuring that the resulting segments capture both the motion patterns and the manipulation actions of the end effector. By integrating these two modalities, we achieve a richer and more accurate segmentation of the policy.

Finally, as a result of the segmentation process, we have a sequence of segments denoted as \( \Sigma_i = \{\sigma_1, \sigma_2, \ldots, \sigma_n\} \), where \( n \) is the number of segments. Here, a segment is expressed as  
  \(
  \sigma = \{s_t, s_{t+1}, \ldots, s_{t+k}\},
  \) comprising $k$ states.


\begin{figure}
	\centering
	\includegraphics[width=\linewidth]{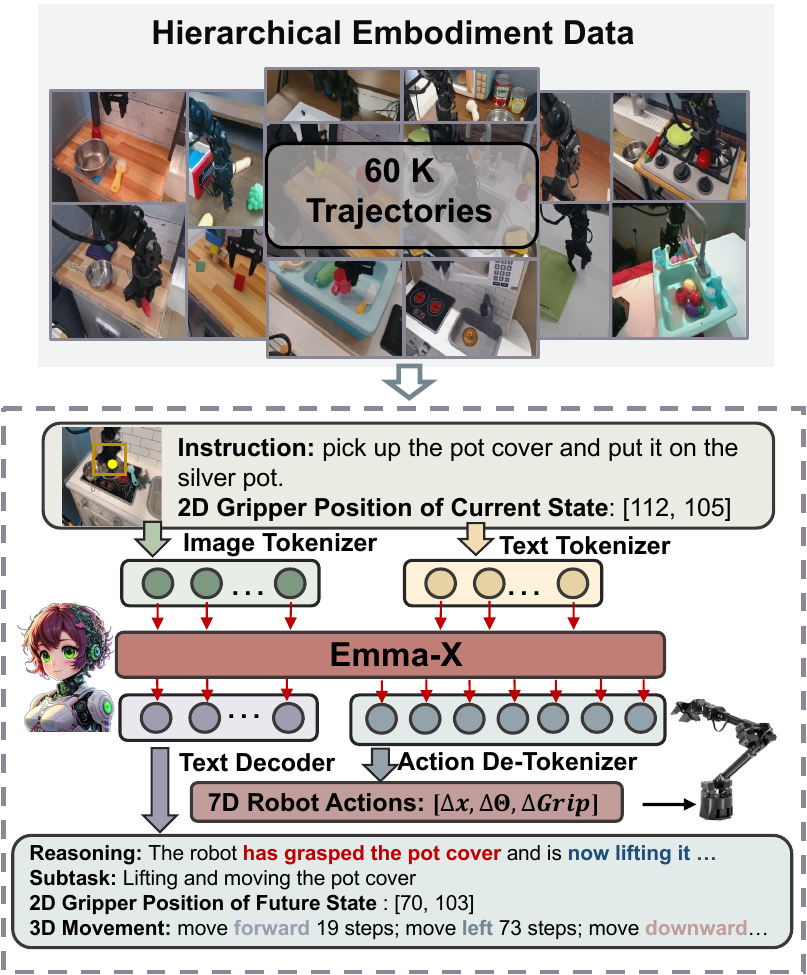}
	\caption{The overview of \model{} fine-tuned from OpenVLA using our hierarchical embodiment dataset.}
	\label{fig:method}
     \vspace{-0.8em}

\end{figure}

\subsection{Data Generation}
After obtaining the segments, we generate hierarchical embodied planning data for each demonstration, as shown in Figure \ref{fig:data}. For each segment of a demonstration, we produce the 2D end-effector position and 3D movements for the completion state of the current segment. Additionally, we generate grounded reasoning for the corresponding subtasks.
\paragraph{Why Look-ahead Spatial Reasoning?}
Consider the robot as a \textit{delivery driver} tasked with delivering a package to a specific destination (the \textbf{goal}). The driver has access to a detailed high-level map of the city, which provides potential landmarks or checkpoints (\( s_{t+k} \)) along the way to the destination. To reach the goal efficiently, the driver performs two tasks:
\textbf{Plans a high-level route}: The driver identifies likely landmarks and routes to guide them toward the destination, akin to predicting \( s_{t+k} \) and the movement plan \(\beta(s_t, s_{t+k})\). \textbf{Executes immediate driving decisions}: While en route, the driver makes real-time decisions (\(a_t\)), such as turning left or stopping at a traffic signal, informed by the planned route and the current position \(s_t\).

Without the ability to establish landmarks or checkpoints (future states) and plan routes based on them, the driver would rely solely on reactive decisions, leading to inefficiencies or incorrect paths. By integrating both the high-level plan and immediate feedback, the driver ensures purposeful and adaptive progress toward the goal. Following this analogy, we calculate the look-ahead gripper position and movement plan to reach there.

\paragraph{Look-ahead Gripper Position Generation.} Following \citep{zawalski2024robotic}, we also use OWLv2 \cite{minderer2024scaling} and SAM \citep{kirillov2023segment} to detect 2D gripper position, which can be seen in Figure \ref{fig:data}(e). The difference is that they train the model to output only the gripper position for the current input state, whereas, in our data construction process, we use the current gripper position as input and predict the gripper position for the first state of the next segment. Thus, although both approaches utilize the gripper position, our model focuses more on predicting the gripper position in future states during training, rather than identifying its position in the current state. Let's consider for every state $s_t$, we obtain $g_t$, the gripper position of the first state of the next segment.
\paragraph{Look-ahead Movement Plan Generation.} 
As shown in Figure \ref{fig:data}(d), we infer the 3D spatial positions corresponding to the current state and the end state of the current segment using the state policy of the robot. Specifically, we calculate the displacement between these two positions to determine the direction and step size required for the manipulator to move from the current state to the end state. Following the motion language idea in RT-H~\cite{belkhale2024rthactionhierarchiesusing}, we encode our high-level motion plans using a standardized template in Appendix \ref{sec:movement-template}. By integrating look-ahead spatial reasoning, the model incorporates both reactive and proactive decision-making. It combines immediate context at the current state \( s_t \) with a high-level plan that predicts likely future states \( s_{t+k} \) and the corresponding movement strategy \( \beta(s_t, s_{t+k}) \). This dual focus enables the model to align immediate actions with the overarching goal, ensuring purposeful and adaptive task execution. Please note that this data is not directly executed as the robot's actions. Let's consider for every state $s_t$, we will obtain $m_t$, the movement plan to the first state of the next segment.

\begin{table*}
\centering
\resizebox{\textwidth}{!}{
\begin{tabular}{l p{7cm} c c c c c c }
\toprule
\multirow{2}*{\textbf{Category}}&\multirow{2}*{\textbf{Task}}&\multicolumn{2}{c}{\textbf{OpenVLA}} & \multicolumn{2}{c}{\textbf{ECoT}} & \multicolumn{2}{c}{\textbf{\model \,(Ours)}}  \\
\cmidrule(r){3-4}
\cmidrule(r){5-6}
\cmidrule(r){7-8}

 && h\_Succ ($\mathbf{\%}$) & Succ ($\mathbf{\%}$) & h\_Succ ($\mathbf{\%}$) & Succ ($\mathbf{\%}$) & h\_Succ ($\mathbf{\%}$) & Succ ($\mathbf{\%}$) \\
\midrule
\SR{}&Put the upper half of the carrot in the pot&30&10&35&20&\textbf{80}&\textbf{60}\\
\SR{}&Put the left half of the lemon in the pan&30 &0 &35 &10 &\textbf{55} &\textbf{20}\\
\SR{} &Put the blue cube on the left plate &25 &20 &5& 0& \textbf{60}& \textbf{60}\\
\SR{}&Put the blue cube on the right plate& 60 &60& 35& 20& \textbf{90}& \textbf{90}\\
\midrule
\OODO{}&Put the banana in pot& 70&50& 45&40 &\textbf{85}&\textbf{70}\\
\OODO{}&Put the blue cube on the plate&\textbf{90} &\textbf{90} &20 &10 &85& 70\\
\OODO{}&Wipe the stove with towel& 70 &50 &50 &30 &\textbf{90} &\textbf{90}\\
\midrule
\OODI{}&Pick up any object that is a kind of vegetable&40&30&15&0&\textbf{75}&\textbf{70}\\
\OODI{}&Put the inedible object on the towel& 0& 0& 25& 0 &\textbf{40}& \textbf{30}\\
\OODI{}&Put the edible object on the towel&0&0&15& \textbf{10}& \textbf{35}& 0\\
\midrule
\ID{}&Open microwave&50 &30& 25&0 &\textbf{65} &\textbf{30}\\
\ID{}&Close microwave&80&60 & 45&40 & \textbf{100} &\textbf{100}\\
\midrule
\multicolumn{2}{l}{\hspace{2.5cm}\textbf{Average}} & 45.41 & 33.33 & 28.75 & 15.00& \textbf{71.66} & \textbf{57.50}\\
\bottomrule
\end{tabular}
}
\caption{Experimental results of \model{} and baselines on 12 real-world WidowX-250 robot manipulation tasks.}
\vspace{-0.8em}
\label{tab:1}
\end{table*}

\paragraph{Grounded Chain-of-Thought Reasoning.} As shown in Figure \ref{fig:data}(f) and (g), we utilize Gemini \footnote{We used gemini-1.5-pro-latest for our data generation.} to derive the subtask corresponding to each segment, along with scene understanding and the reasoning behind the series of actions the robot needs to perform the subtask. Specifically, we take sequences of segmented images, and task descriptions as input to guide Gemini in generating the subtask and grounded reasoning for each segment. Compared to \citep{zawalski2024robotic} that infer subtasks and their mapping to states solely from textual information, our approach first segments the sequence based on the robot's motion trajectory and gripper's state as explained in Section \ref{sec:segment}. After that, based on the given multimodal information, we generate the corresponding subtasks and the reasoning of each subtask. Note that each subtask can comprise multiple segments. For the \(i\)-th trajectory, we obtain the grounded reasoning from Gemini, defined as:  
\(
GR_i = \big\{ (\sigma_k, \mathcal{S}_k, \mathcal{R}_k) \mid k = 1, \ldots, n \big\},
\)  
where:  
- \(\sigma_k\) is the \(k\)-th segment,  
- \(\mathcal{S}_k\) is the subtask label assigned to \(\sigma_k\),  
- \(\mathcal{R}_k\) is Gemini's justification for assigning subtask \(\mathcal{S}_k\) to \(\sigma_k\), and  
- \(n\) is the total number of segments in the trajectory.
The prompt template can be seen in the Appendix \ref{sec:prompt-template}.

\paragraph{The Final Dataset.}  
The final dataset for the \(i\)-th trajectory in the training dataset is defined as:  
\(
\{\mathcal{D}_i\}_{i=1}^{N} = \{X_i, Y_i\}_{i=1}^{N} = \big\{ \{(s_t, \mathcal{T}_i), (m_t, g_t, GR_t, a_t)\}_{t=1}^{T}\big\}_{i=1}^{N},
\) 
where \(t = 1, 2, \ldots, T\), and \(T\) is the total number of timesteps in the trajectory.

\subsection{\model{}}

In this section, we introduce the architecture of our proposed \model \, which is a 7B-parameter VLA model fine-tuned from OpenVLA using our constructed hierarchical embodiment data. As shown in Figure \ref{fig:method}, we adjust the text prompt with the current gripper position and add chain-of-thought training to enhance the ability of spatial reasoning and scene understanding before predicting the next robot action policy. 

During the process of predicting for real robot testing, we input the task description, the current observation image, and the 2D gripper position detected in real-time by OWLv2 \cite{minderer2024scaling} and SAM \citep{kirillov2023segment}. \model \, first outputs the subtask and a description of the current scene, including the spatial relationship between the target object in the image and the robotic arm, as well as the operational instructions required for the gripper to reach the goal of the current subtask. Additionally, \model \, also predicts the target position the gripper needs to reach after completing the sub-task, including both the 2D location in the image and the 3D spatial movements. Finally, the model outputs the next 7D robot action policy for downstream manipulation.

\section{Experiments}
\subsection{Implementation Details}

To create the hierarchical reasoning dataset, we employed our data creation pipeline on full BridgeData-v2, which consists of approximately 60,000 trajectories paired with task instructions, resulting in an augmented dataset.

To train our VLA models, we employed OpenVLA, a 7B vision-language-action (VLA) model built upon the Prismatic vision-language framework and pretrained on the Open X-Embodiment dataset. For autoregressive training, we tokenized our 7-dimensional action policy into discrete policy tokens, consistent with OpenVLA's methodology. We adhered to OpenVLA's training procedure and fine-tuned the base model on our augmented dataset for 3 epochs until convergence.

\begin{figure}[t!]
        \centering
         \vspace{-0.8em}
        \begin{tikzpicture}
    \definecolor{customRed}{HTML}{F5867F}
    \definecolor{customYellow}{HTML}{FFBC80}
    \definecolor{customBlue}{HTML}{6B98C4}
    \definecolor{customPurple}{HTML}{EEBEC0}
        \begin{axis}[
            ybar,
            bar width=.3cm,
            width=0.48\textwidth,
            height=4.5cm,
            enlarge x limits=0.15,
            ylabel={Avg h\_Succe Rate ($\mathbf{\%}$) },
            xlabel={Different Categories of Tasks},
            symbolic x coords={ \SR, \OODO, \ID, \OODI},
            xtick=data,
            ymin=0,
            ymax=100,
            ytick={0,10,20,30,40,50,60,70,80,90,100},
            grid=major,
            xmajorgrids=false, 
            tick label style={
        font=\fontsize{6}{1}\selectfont 
    },
    xlabel style={
        font=\fontsize{9}{1}\selectfont 
    },
    ylabel style={
        font=\fontsize{9}{1}\selectfont 
    },
            grid style={dashed,gray!30},
                    legend style={
                font=\fontsize{7}{1}\selectfont, 
                legend style={row sep=-0.1cm},
                at={(1,1)},
            },
            legend image code/.code={
              \draw[#1] (0cm,-0.1cm) rectangle (0.15cm,0.1cm);
            }, 
        ]
        
        \addplot [fill=customRed] coordinates {(\SR,28) (\OODO,38)  (\ID, 15)  (\OODI,18)};
        \addplot [fill=customYellow] coordinates {(\SR,36) (\OODO,77) (\ID,65) (\OODI,13) };
        \addplot [fill=customBlue] coordinates {(\SR,71) (\OODO,87)  (\ID,83)(\OODI,50)};
        \legend{ECoT,OpenVLA,  \model}   
        \end{axis}
    \end{tikzpicture}
    \vspace{-1.2em}
    \caption{Experimental results on different categories of real-world robot tasks.}
    \vspace{-0.8em}
    \label{fig:lengths}
\end{figure}
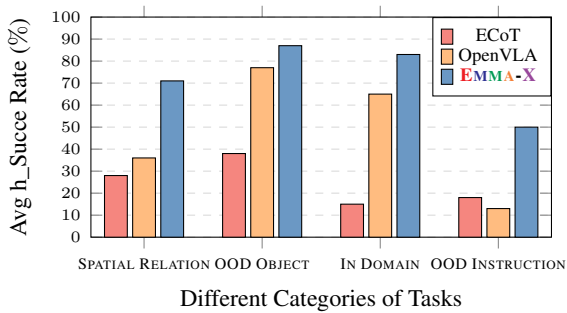

\subsection{Robot Setup and Metrics}
We evaluate our approach using the 6-DoF WidowX robot arm, as introduced in the Bridge V2 paper, which represents a standard benchmark for assessing generalizable robotic policies. The policy takes as input a single third-person camera feed and a natural language instruction, predicting end-effector velocity actions to control the robot.

To rigorously test the generalization capabilities of the policies, we develop a suite of challenging evaluation tasks that span multiple aspects: in-domain scenarios, out-of-domain (OOD) objects, spatial relationships, and OOD instructions. All policies are assessed on identical real-world setups to ensure consistency in camera angle, lighting conditions, and background. Each task is conducted over $10$ trials, adhering to the methodology established by OpenVLA. If the robot can successfully achieve the task specified inside the prompts, it is counted as a success (\textbf{succ}) receiving a score of $1$, otherwise, a score of $0$ is assigned. Following OpenVLA, we also introduce a "half-success" (\textbf{h-succ}) metric that considers both the task goal and difficulty and assigns a $0.5$ score only when the half-success criteria are met (Appendix \ref{sec:half-success}).

\subsection{Baselines}
To comprehensively evaluate the performance of our proposed \model, we conduct extensive experiments across 12 different tasks on the real robot with several competitive methods.


\textbf{OpenVLA} \citep{kim2024openvla}: A VLA model based on large-scale VLM Prismatic-7b and pre-trained on the Open-X-Embodiment dataset\citep{open_x_embodiment_rt_x_2023}.

\textbf{OpenVLA w/ FT}: For a fair comparison, we finetuned the OpenVLA model on the BridgeV2 dataset for the same number of epochs following the same training setting in our method.

\textbf{ECoT} \citep{zawalski2024robotic}: A VLA model fine-tuned from OpenVLA on BridgeV2 dataset \citep{walke2023bridgedata} with their generated chain-of-thought reasoning data.

\begin{figure*}
\vspace{-0.8em}
	\centering
	\includegraphics[width=\textwidth]{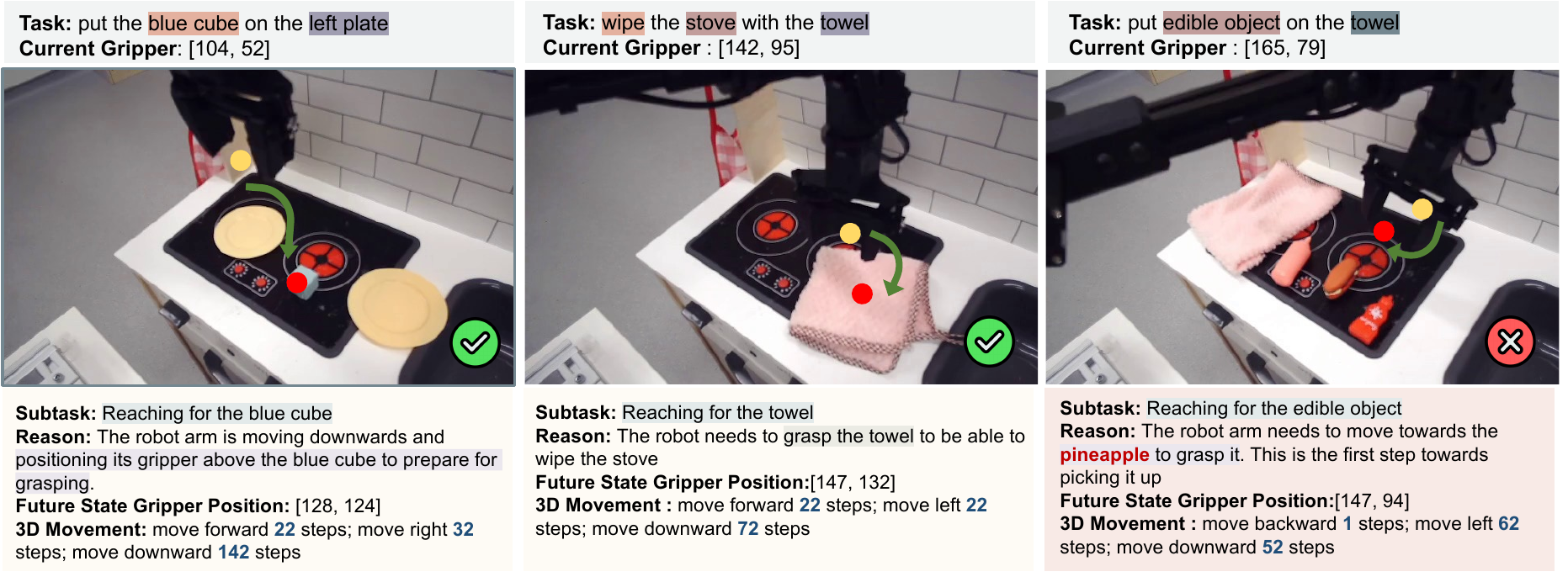}
	\caption{Qualitative examples of successful and failed cases with \model{} on real-world robot testing.}
	\label{fig:case}
     \vspace{-0.8em}

\end{figure*}



\begin{table}
\centering
\setlength{\tabcolsep}{3.5pt} 
\resizebox{\linewidth}{!}{
\begin{tabular}{l l l l l l l l}
\toprule
\multirow{2}*{\textbf{Models}} & \multicolumn{2}{c}{\textbf{\SR{}}} & \multicolumn{2}{c}{\textbf{\OODO{}}} & \multicolumn{2}{c}{\textbf{\OODI{}}} \\
\cmidrule(r){2-3} \cmidrule(r){4-5} \cmidrule(r){6-7}
 & h\_Succ (\%) & Succ (\%) & h\_Succ (\%) & Succ (\%) & h\_Succ (\%) & Succ (\%) \\
\midrule
\model & 77 & 70 & 88 & 80 & 75 & 70 \\
\hspace{2mm}w/o \(m_t\) & 42 \textcolor{red}{(\(\downarrow 35\))} & 37 \textcolor{red}{(\(\downarrow 33\))} & 63 \textcolor{red}{(\(\downarrow 25\))} & 55\textcolor{red}{(\(\downarrow 25\))}  & 40 \textcolor{red}{(\(\downarrow 35\))} & 30 \textcolor{red}{(\(\downarrow 40\))} \\
\hspace{2mm}w/o \(g_t\) & 32 \textcolor{red}{(\(\downarrow 45\))}  & 30 \textcolor{red}{(\(\downarrow 40\))} & 45 \textcolor{red}{(\(\downarrow 43\))} & 35 \textcolor{red}{(\(\downarrow 45\))}  & 45 \textcolor{red}{(\(\downarrow 30\))} & 30 \textcolor{red}{(\(\downarrow 40\))}\\
\hspace{2mm}w/o \(GR_t\) & 22 \textcolor{red}{(\(\downarrow 55\))} & 10 \textcolor{red}{(\(\downarrow 60\))} & 45 \textcolor{red}{(\(\downarrow 43\))} & 40 \textcolor{red}{(\(\downarrow 40\))} & 25 \textcolor{red}{(\(\downarrow 50\))} & 20 \textcolor{red}{(\(\downarrow 50\))} \\
\hspace{2mm}w/o HDBSCAN & 27 \textcolor{red}{(\(\downarrow 50\))} & 20 \textcolor{red}{(\(\downarrow 50\))} & 53 \textcolor{red}{(\(\downarrow 35\))} & 35 \textcolor{red}{(\(\downarrow 45\))} & 65 \textcolor{red}{(\(\downarrow 10\))} & 40 \textcolor{red}{(\(\downarrow 30\))} \\
\midrule
OpenVLA & 38 & 30 & 70 & 50 & 40 & 30 \\
\hspace{2mm}w/ FT & 28 \textcolor{red}{(\(\downarrow 10\))} & 23 \textcolor{red}{(\(\downarrow 17\))} & 65 \textcolor{red}{(\(\downarrow 5\))} & 50 \textcolor{red}{(\(\downarrow 0\))} & 15 \textcolor{red}{(\(\downarrow 25\))} & 0 \textcolor{red}{(\(\downarrow 30\))}\\
\bottomrule
\end{tabular}
}
\caption{\label{tab:2} Models with different configurations.}
\vspace{-0.8em}
\end{table}

\subsection{\model{} Improves Policy Generalization}

In this section, we compare \model \, with several baselines on 12 real-world robotic tasks. As shown in Table \ref{tab:1}, our \model \, outperforms the strong baseline OpenVLA, with a 24.17\% increase in task success rate and a 26.25\% increase in half success rate. This demonstrates the effectiveness of our constructed hierarchical embodiment dataset. In addition, compared to ECOT, our \model \, shows significant gains, which can be caused by the following: 1) ECoT suffers from noisy training data, which causes hallucinations when faced with out-of-domain instructions or unfamiliar objects, leading to task failures. Interestingly, even for \ID{} tasks, it performs poorly compared to other models, highlighting its limited reasoning capabilities.
Our grounded task reasoning approach addresses this by incorporating the segmented visual images, ensuring more accurate task understanding. 2) \model \, enhances spatial reasoning by predicting the 2D gripper position of the end state of the current segment and 3D spatial movements to transit to it before predicting the next robot action policy.

As shown in Figure \ref{fig:lengths}, we also compared the average performance across various categories of robotic tasks. Notably, our method achieved the most significant performance improvement in \SR{} tasks, outperforming OpenVLA by 35\% and ECoT by 29\% in the h\_Succ rate. These results strongly validate the effectiveness of our predicted 3D spatial movements. Furthermore, our method demonstrated substantial performance gains in \OODI{} tasks, highlighting the efficacy of our grounded task reasoning.

\subsection{Analysis}
We trained several variants of \model{} to evaluate the roles of segmentation, look-ahead spatial reasoning, and grounded chain-of-thought (CoT) reasoning, which collectively constitute the core of \model{}. For this evaluation, we sampled 6 prompts across  \SR{}, \OODO{}, and \OODI{} (prompts are indicated in \textcolor{magenta}{magenta} color in Section \ref{sec:half-success}). For each prompt, we conducted 10 rollouts under the same experimental setup as our main experiments.
\paragraph{Segmentation Greatly Helps the Policy.}
To evaluate the effectiveness of our segmentation technique, we conducted an experiment where sequences were segmented solely based on the gripper's (end effector) open and close positions. The results, reported in Table \ref{tab:2} under the \textit{w/o HDBSCAN} condition, show a general performance drop of 10\% to 50\%. Notably, spatial reasoning performance experienced the most significant decline, with a drop of 50\%. These findings demonstrate that the distance metric introduced in Eq.~\ref{eq:distance-measurement} is crucial for the segmentation process. 

\paragraph{The Impact of Look-ahead Spatial Reasoning.}
To evaluate the importance of look-ahead spatial reasoning, we conducted two experiments:  
1) \model{} was trained without explicitly predicting the gripper's position in the next segment, relying only on the predicted movement plan to reach the future gripper position of that segment (denoted as \textit{w/o \(g_t\)} in Table \ref{tab:2}). This assumes that \model{} implicitly infers the future gripper position.  
2) We trained \model{} to predict the future end effector's position but without rolling out a movement plan to reach that position (denoted as \textit{w/o \(m_t\)} in Table \ref{tab:2}). The results reveal significant performance drops in both cases (25\%-40\% for ``w/o \(m_t\)'' and 30\% to 45\% for ``w/o \(g_t\)''), with a more pronounced decline in spatial reasoning tasks (35\% for ``w/o \(m_t\)'' and 45\% for ``w/o \(g_t\)''). Furthermore, the results suggest that predicting the future end effector's position is more critical, as the performance drop in the absence of 3D spatial movements to the next segment is less severe. We hypothesize that this may be due to OpenVLA's inherent spatial reasoning capabilities, which enable it to more easily transition between positions.
\paragraph{The Importance of Grounded CoT Reasoning.}
Grounded chain-of-thought (CoT) reasoning is a foundational element of \model{}. To assess its impact, we trained a variant of \model{} without grounded reasoning, while retaining look-ahead spatial reasoning in the data. The results show a marked performance drop by 43\%-55\%, highlighting that spatial reasoning alone is insufficient. Interestingly, the absence of grounded CoT reasoning resulted in a more severe decline in spatial reasoning performance compared to models where spatial reasoning capabilities were explicitly ablated. This underscores the critical role of grounded CoT in tackling complex reasoning tasks, including spatial reasoning. Therefore, we surmise that for enhancing the generalizable policies of Vision-Language-Action (VLA) models, it is essential to improve their broader reasoning capabilities, encompassing object recognition, color understanding, abstraction, commonsense knowledge, and more.

\paragraph{Fine-tuning does not Improve OpenVLA.}
We sought to find whether fine-tuning OpenVLA on BridgeV2 could match the performance of \model{}. The results, shown in Table \ref{tab:2}, reveal that OpenVLA's performance degrades by 5\%-30\% after fine-tuning with the worst performance observed for \OODI. We hypothesize that this decline is due to overfitting, as BridgeV2 is already part of OpenVLA's pre-training dataset.

\paragraph{Qualitative Analysis on Real-world Robot Task.}
To qualitatively evaluate the effectiveness of our spatial and task reasoning in guiding robotic actions, we present two successful trajectories and one failed trajectory in Figure \ref{fig:case}. From the left case, we find that the predicted gripper position corresponds to the end state of the subtask ``reaching for the blue cube''. The 3D movement provides a detailed path, clearly directed toward the ``blue cube''. We also include a failed trajectory where the ``hotdog" is mistakenly identified as a ``pineapple". This error propagates, impacting the prediction of the gripper's future position and preventing it from accurately picking up the ``hot dog''.

\section{Conclusion}

We introduce \model{}, a 7B-parameter embodied multimodal action model designed to enhance spatial reasoning and task planning for robotic policy generation. We construct a hierarchical embodiment dataset enriched with grounded reasoning, including 2D gripper positions and 3D spatial movements. Furthermore, our proposed trajectory segmentation strategy reduces hallucination in task reasoning by grounding reasoning in visual images. The experimental results demonstrate the effectiveness of \model{}, showing significant improvements over existing baselines in tasks requiring long-horizon spatial reasoning. 

\section*{Limitations}
While \model{} shows promising performance, its latency remains higher compared to OpenVLA. This increased inference time primarily results from the additional tokens generated during the reasoning process. Specifically, \model{} generates approximately 10 times more tokens than OpenVLA. To mitigate this, a potential strategy is to predict all policies within a segment and only regenerate the policy if the predicted policy deviates significantly from the expected movement plan.
Another limitation is the generalization capability of \model{}. Scaling the training process to incorporate a larger subset of the OXE dataset could enhance the model's ability to handle a broader range of tasks and robotic systems.
Lastly, using SAM for detecting the gripper position can lead to inaccuracies. These errors may occur when the gripper is partially occluded by objects or positioned outside the image frame. Employing a more robust model for detecting and segmenting the robot hand could address these challenges and improve reliability.
%

\bibliography{custom}

\appendix
\onecolumn
\label{sec:appendix}

\section{Related Work} \label{app:related_works}
\paragraph{Generalist Robot Policies.}
Recent progress in robotics has shifted focus towards developing multi-task "generalist" robot policies capable of handling a wide variety of tasks across diverse robot embodiments \citep{brohan2023rt1roboticstransformerrealworld,brohan2023rt2visionlanguageactionmodelstransfer,ebert2021bridgedataboostinggeneralization,walke2023bridgedata,open_x_embodiment_rt_x_2023,octo_2023,kim2024openvla}.
For example, Octo \citep{octo_2023} utilizes a compositional design to train a generalist policy capable of handling various tasks directly while supporting fine-tuning for new inputs and action spaces. 
Similarly, OpenVLA \citep{kim2024openvla} adopts a streamlined end-to-end approach, fine-tuning vision-language models (VLMs) to produce robot actions by treating these actions as tokens within the language model's vocabulary.
These studies highlight the potential of training robot policies on large and diverse datasets as a promising strategy for enhancing their performance.

\paragraph{Vision-Language-Action Models.}
A number of recent works have explored fine-tuning large pretrained VLMs for predicting robot actions \citep{open_x_embodiment_rt_x_2023,brohan2023rt2visionlanguageactionmodelstransfer,kim2024openvla,octo_2023,driess2023palmeembodiedmultimodallanguage}
Such models are often referred to as vision-language-action models (VLAs), as they fuse robot actions directly into VLM backbones and treating these actions as tokens within the language model vocabulary.
This approach provides a simple yet scalable alternative, with models such as RT-2 \citep{brohan2023rt1roboticstransformerrealworld}, RT-2-X \citep{open_x_embodiment_rt_x_2023}, and OpenVLA \citep{kim2024openvla} demonstrating state-of-the-art performance and impressive generalization across diverse objects and environments.
RT-2 integrates Internet-scale vision-language data with robotic trajectory data, while RT-2-X scales this further with a 55B-parameter policy trained on the Open X-Embodiment dataset \citep{open_x_embodiment_rt_x_2023}.
In contrast, OpenVLA integrates a robust open VLM backbone with an enriched robot pretraining dataset.
Despite these advancements, current VLAs underutilize some of the most valuable features of their underlying language and vision-language models, specifically, their capacity to reason through the steps needed to solve complex tasks.

\paragraph{Reasoning for Robotics.}
Prompting large language models (LLMs) to "think step-by-step" \citep{kojima2022large} when solving problems can significantly enhance their performance. 
Similar techniques have been explored in the context of high-level task planning for robotics \citep{liang2023codepolicieslanguagemodel,ha2023scalingdistillingdownlanguageguided}. 
Expanding on this, \citet{zawalski2024robotic} introduced ECoT, a method that trains a VLA policy to autoregressively generate chain-of-thought (CoT) reasoning. 
ECoT combines high- and low-level reasoning with actionable steps, aligning these to an agent's environment based on input instructions and observations. 
While this equips VLAs with a better understanding of the current situation and task, it falls short in two key areas: long-horizon spatial reasoning for robot movement and visual grounding when augmenting reasoning data with Gemini.
Building on these limitations, we build \model{} designed to enhance spatial reasoning and task planning in robotic policy generation. \model{} integrates grounded chain-of-thought reasoning to predict future states and transitions, enabling robots to develop more informed and effective action policies.

\section{Prompts and Sample Outputs}
\label{sec:prompt-template}
Template for prompting Gemini to generate subtasks and reasoning.

\begin{lstlisting}
The robot successfully completed a task specified by the instruction: '{instruction}'. Here is a list of segments of images for the robot hand to perform the task specified by the instruction.

Based on the sequence of segments provided in sequential order, pay attention to the robot hand and identify which subtask it is performing in each segment. Provide the justification for why the subtask should be done based on the environment. You can assign the same subtask to multiple segments.

You should output in dictionary format: {segment_number: [subtask, reason for justification], ...}. The segment_number starts from 1 and must be an integer. The output dictionary keys correspond to each segment, and the output dictionary length should be the same as the number of segments.

{segment_count}

Segment 1: images in segment 1  
Segment 2: images in segment 2  
...  
Segment n: images in segment n
\end{lstlisting}

We list the Gemini output of the example in Figure \ref{fig:intro} as follows:
\begin{lstlisting}
Here\'s a breakdown of the robot\'s subtasks in each segment:
   
```json{
   1: ["Positioning over pot cover", "The robot arm needs to move to the correct location above the pot cover to be able to grasp it. This involves adjusting its position and orientation."],
   2: ["Grasping the pot cover", "The robot is closing its gripper around the pot cover\'s handle to securely pick it up. This requires precision to ensure a stable grip."],
   3: ["Lifting and moving the pot cover", "The robot has grasped the pot cover and is now lifting it up and moving it towards the silver pot.  It must avoid collision with other objects in the environment."],
   4: ["Positioning and placing the cover on the pot", "The robot carefully positions the cover above the silver pot and lowers it into place. This requires aligning the cover with the pot\'s opening and releasing the grip gently to avoid dropping or damaging the cover or pot."]
   }
   
\end{lstlisting}

\section{Half Success Criteria}
\label{sec:half-success}
The half-success criteria to for the metric h\_success are as follows. The prompts in \textcolor{magenta}{magenta} color are used to report the results in Table \ref{tab:2}.

\begin{tabular}{|p{7cm}|p{7cm}|}
\hline
\textbf{Task} & \textbf{Half Success Criteria} \\
\hline
\textcolor{magenta}{Put the banana in pot} & Robot successfully grasps the banana. \\ \hline
\textcolor{magenta}{Put the upper half of the carrot in the pot} & Robot moves towards or grasps the correct target object. \\ \hline
\textcolor{magenta}{Pick up any object that is a kind of vegetable} & Robot moves towards the correct vegetable. \\ \hline
Put the left half of the lemon in the pan & Robot moves towards and correctly selects the left half of the lemon. \\ \hline
Put the blue cube on the plate & Robot grasps the blue cube. \\ \hline
\textcolor{magenta}{Put the blue cube on the left plate} & Robot grasps the blue cube and moves it towards the left plate. \\ \hline
\textcolor{magenta}{Put the blue cube on the right plate} & Robot grasps the blue cube and moves it towards the right plate. \\ \hline
Put the inedible object on the towel & Robot moves towards the correct inedible object. \\ \hline
Put the edible object on the towel & Robot moves towards the correct edible object. \\ \hline
\textcolor{magenta}{Wipe the stove with a towel} & Robot touches the towel but does not wipe the stove. \\ \hline
Open the microwave & Robot partially opens the microwave door. \\ \hline
Close the microwave & Robot partially closes the microwave door. \\ \hline
\end{tabular}
\section{Segmentation Statistics}
Average Number of frames per segment: 5.5 \\
Average Number of segments per trajectory: 6.9 \\
Average Number of frames per trajectory: 32.8 \\

\section{Motion Plan Template}
\label{sec:movement-template}
\textbf{Translational Movements:} \textit{move (left/right)} $x$ steps, \textit{move (forward/backward)} $y$ steps, \textit{move (upward/downward)} $z$ steps.\\
\textbf{Rotational Movements:} \textit{pitch (upward/downward)} $\alpha$ degrees, \textit{yaw (left/right)} $\beta$ degrees, \textit{roll (clockwise/counterclockwise)} $\gamma$ degrees.\\
\textbf{Gripper Action:} \textit{(open/close)} gripper.

\section{Pseudo Code for Training \model{} and Running Inference}
Notations defined in \ref{ss:hierachical_policy_imitation}

\begin{algorithm}
\caption{Data Generation, Training, and Inference Process}
\label{alg:full_process}

\textbf{For each sample $i$ in Embodied Dataset, we have:} \\
\hspace{0.5cm} $T$: Number of time frames \\
\hspace{0.5cm} $\mathbf{S} = \{s_t\}_{t=1}^{T}$: Images at each time frame $t$, where $s_t$ is the image at time frame $t$ \\
\hspace{0.5cm} $\mathbf{G} = \{g_t\}_{t=1}^{T}$: Gripper poses at each time frame $t$ (position, orientation, and open-or-close state) \\
\hspace{0.5cm} $\mathcal{T}$: Task instruction in natural language format \\

\vspace{0.2cm}
\textbf{Training Process:}

\begin{algorithmic}[1]
\While{not converged}
    \For{each sample $i$}
            \State $M_{\text{frames} \to \text{segments}} \gets \text{dual\_segmentation}(G)$ \Comment{Mapping from frame to segment}
            \State $M_{\text{segments} \to \text{subtasks, reasons}} \gets \text{Gemini}(S, G, \mathcal{T})$ \Comment{Mapping from segment to subtasks}
        \For{each time frame $t \in \{1, 2, \ldots, T\}$}

            \State $\sigma_t \gets M_{\text{frames} \to \text{segments}}(t)$ \Comment{Get segment for time $t$}
            \State $\mathcal{GR}_t \gets M_{\text{segments} \to \text{subtasks, reasons}}(\sigma_t)$ \Comment{Get grounded reasoning from Gemini}
            
            \State $g_t \gets \text{SAM}(s_t)$ \Comment{Get 2D gripper position Using SAM model}
            \State $g_{\text{end}} \gets \text{SAM}(S_{\text{end}})$ \Comment{Get 2D gripper position at end of current segment}
            
            \State $m_t \gets \text{Template}(g_t - g_{\text{end}})$ \Comment{Translational change to movement plan in natural language}
            
            \State $prediction \gets \text{Model}(\mathcal{T}, s_t, g_t, GR_t, g_{\text{end}}, m_t, a_t)$ \Comment{Perform supervised fine-tuning (SFT) with label: ($\mathcal{T}, s_t, g_t, GR_t, g_{\text{end}}, m_t, a_t$) }
        \EndFor
    \EndFor
\EndWhile
\end{algorithmic}

\vspace{0.2cm}
\textbf{Inference Process:}

\begin{algorithmic}[1]
\While{Task not completed}
    \State $g_t \gets \text{SAM}(s_t)$
    \State $GR_t, g_{\text{end}}, m_t, a_t \gets$ \hspace{0.5cm} \model{}($\mathcal{T}, s_t, g_t$)
    \State Control the robot using $a_t$, to get new $s_t, g_t$
\EndWhile
\end{algorithmic}

\end{algorithm}

\end{document}